\pdfoutput=1

\documentclass[11pt]{article}

\usepackage{ACL2023}

\usepackage{times}
\usepackage{latexsym}

\usepackage[T1]{fontenc}

\usepackage[utf8]{inputenc}

\usepackage{microtype}

\usepackage{inconsolata}
\usepackage{CJKutf8}
\usepackage{caption}
\usepackage{bm}
\usepackage{amsmath}
\usepackage{color}
\usepackage{graphicx}
\usepackage{subfigure}
\usepackage{ragged2e}
\usepackage{booktabs}
\usepackage{arydshln}
\usepackage{multirow}
\usepackage{amsfonts,amssymb}
\title{Type-enhanced Ensemble Triple Representation via Triple-aware Attention for Cross-lingual Entity Alignment}

 \author{Zhishuo Zhang \and Chengxiang Tan\thanks{Corresponding author.} \and Haihang Wang \and Xueyan Zhao \and Min Yang \\
 Department of Computer Science and Technology, Tongji University, Shanghai, China \\
         \texttt{\{2110140,jerrytan,wanghh,1710839,2110136\}@tongji.edu.cn}
     }
\begin{document}
\begin{sloppypar}
\begin{CJK}{UTF8}{gbsn}
\maketitle
\begin{abstract}
Entity alignment(EA) is a crucial task for integrating cross-lingual and cross-domain knowledge graphs(KGs), which aims to discover entities referring to the same real-world object from different KGs.
Most existing methods generate aligning entity representation by mining the relevance of triple elements via embedding-based methods, paying little attention to triple indivisibility and entity role diversity.
In this paper, a novel framework named TTEA -- \textbf{T}ype-enhanced Ensemble Triple Representation via \textbf{T}riple-aware Attention for Cross-lingual \textbf{E}ntity \textbf{A}lignment is proposed to overcome the above issues considering ensemble triple specificity and entity role features.
Specifically, the ensemble triple representation is derived by regarding relation as information carrier between semantic space and type space, and hence the noise influence during spatial transformation and information propagation can be smoothly controlled via specificity-aware triple attention. Moreover, our framework uses triple-ware entity enhancement to model the role diversity of triple elements.
Extensive experiments on three real-world cross-lingual datasets demonstrate that our framework outperforms state-of-the-art methods.
\end{abstract}

\section{Introduction}
Cross-lingual knowledge graphs(KGs) such as DBpedia\citep{bizer_dbpedia_2009}, YAGO\citep{suchanek_yago_2008} and ConceptNet\citep{speer_conceptnet_2018} have been widely applied in many real-world scenarios, such as finance\citep{financial}, medical care\citep{medical1,Medical}, and artificial intelligence\citep{2020Open,recommand,question}. As most KGs are independently constructed in different languages or domains, data formatted as $(head\ entity,relation,tail\ entity)$ cannot be effectively integrated due to heterogeneity and rule specificity.
Entity alignment(EA) is a crucial task for information fusion, which aims to discover equivalent entities from different KGs.

Recently, embedding-based methods have attracted wide attention, which embed entity and relation by encoding them into latent vector spaces and measure the embedding distance for EA\citep{wang_cross-lingual_2018, mao_relational_2020, peng_embedding-based_2020, zhang_adaptive_2021}. There have been many efforts to obtain excellent representation of entity and relation for EA, which can be roughly divided into two categories according to motivations: Trans-based methods and GNNs-based methods.

\begin{figure}[t]
	\centering
	\includegraphics[scale=.46]{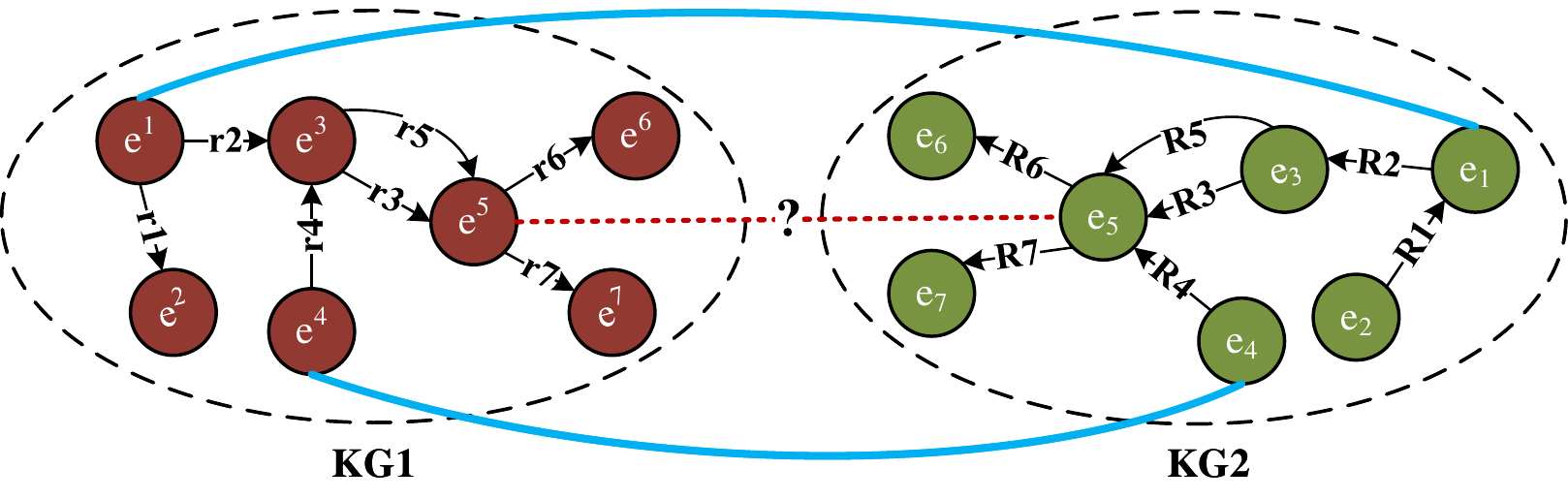}
	\caption{A toy example of entity-role diversity and multiple relations between entities. The blue solid lines between entities in {KG1} and {KG2} refer to the alignments.}
	\label{FIG:1}
\end{figure}

Trans-based methods treat element interaction as a translation process $\boldsymbol{h\!+\!r\! \approx \!t}$ for a triple $\boldsymbol{(h,r,t)}$. These methods\citep{transh2014, lin_learning_2015, sun_transedge_2019} are effective and simple, but unable to form the complete representation of triple elements as the internal correlation is complex and indescribable. GNNs-based methods fall into one of two categories, GCNs-based and GATs-based. The former usually reflect EA via neighbor alignment and topology structure \citep{gao_mhgcn_2022, dual_gated, zhu_relation-aware_2020}, and the latter integrate the surrounding information to enhance embedding\citep{wu_relation-aware_2019, zhu_raga_2021}. Although these methods can effectively improve performance via fusing neighbor information, they rarely consider the specificity of ensemble triple and role diversity. As depicted in Figure \ref{FIG:1}, given the aligned pairs {$\left(e^{1}, e_{1}\right)$}, the entity {$e^{1}$} plays a role as head entity in {KG1} and {$e_{1}$} as tail entity in {KG2}, it is intuitive that the influence of {$e^{1}$} on triple {$\left(e^{1}, r1 ,e^{2}\right)$} is inconsistent with the influence of {$e_{1}$} on triple {$\left(e_{2}, R1 ,e_{1}\right)$}. Furthermore, there may be multiple relations holding different types between head entity and tail entity, as the entity pairs {$\left(e^{3}, e^{5}\right)$} in {KG1} and {$\left(e_{3}, e_{5}\right)$} in {KG2} show.

To address the above shortcomings, we propose TTEA -- \textbf{T}ype-enhanced Ensemble Triple Representation via \textbf{T}riple-aware Attention for Cross-lingual \textbf{E}ntity \textbf{A}lignment in this paper with the intuitive assumption that relation is generic in semantic space and type space for a specific triple, which is capable of fully utilizing triple specificity and role diversity. Considering that triple elements are indivisible, TTEA introduces a type-enhanced ensemble triple representation module to capture semantic and type information while preserving triple specificity. In terms of multiple relations and roles in a triple, we design a triple-aware entity enhancement mechanism to obtain cycle co-enhanced head-aware and tail-aware entity embedding. To our best knowledge, TTEA is the first work to exploit ensemble triple specificity and role diversity of head and tail entities for EA. Experimental results on three cross-lingual KGs prove that TTEA outperforms state-of-the-art baselines. The source code is available in github\footnote{\url{https://github.com/CodesForNlp/TTEA}}.

In summary, our main contributions are as follows:

$\bullet$We provide a novel perspective to regard relation as information carrier during spatial transformation, which is capable to effectively alleviate the noise introduced during mapping.

$\bullet$ We propose a novel EA framework which sufficiently utilizes triple specificity and role diversity via ensemble triple representation and triple-aware entity enhancement.

$\bullet$ Extensive experiments conducted on public datasets demonstrate that TTEA significantly and consistently outperforms state-of-the-art EA baseline methods.
\section{Problem Formulation}\label{sec3}
A KG could be formalized as $KG = (E, R, T)$, where $E$ and $R$ are the sets of entity and relation respectively, $T \subset E \times R \times E$ is the set of relational triple. Given two cross-lingual KGs, $KG1=(E1, R1, T1)$ and $KG2=(E2, R2, T2)$, the task of entity alignment is defined as discovering entity pairs referring to the same real-world object in $KG1$ and $KG2$ based on a set of seed entity pairs, which is denoted as $S=\{(e_1, e_2)|e_1\in E1, e_2 \in E2\}$, where $e_1$ and $e_2$ are equivalent.
\section{TTEA Framework}\label{sec4}
\begin{figure*}[htbp]
	\centering
	\includegraphics[scale=.77]{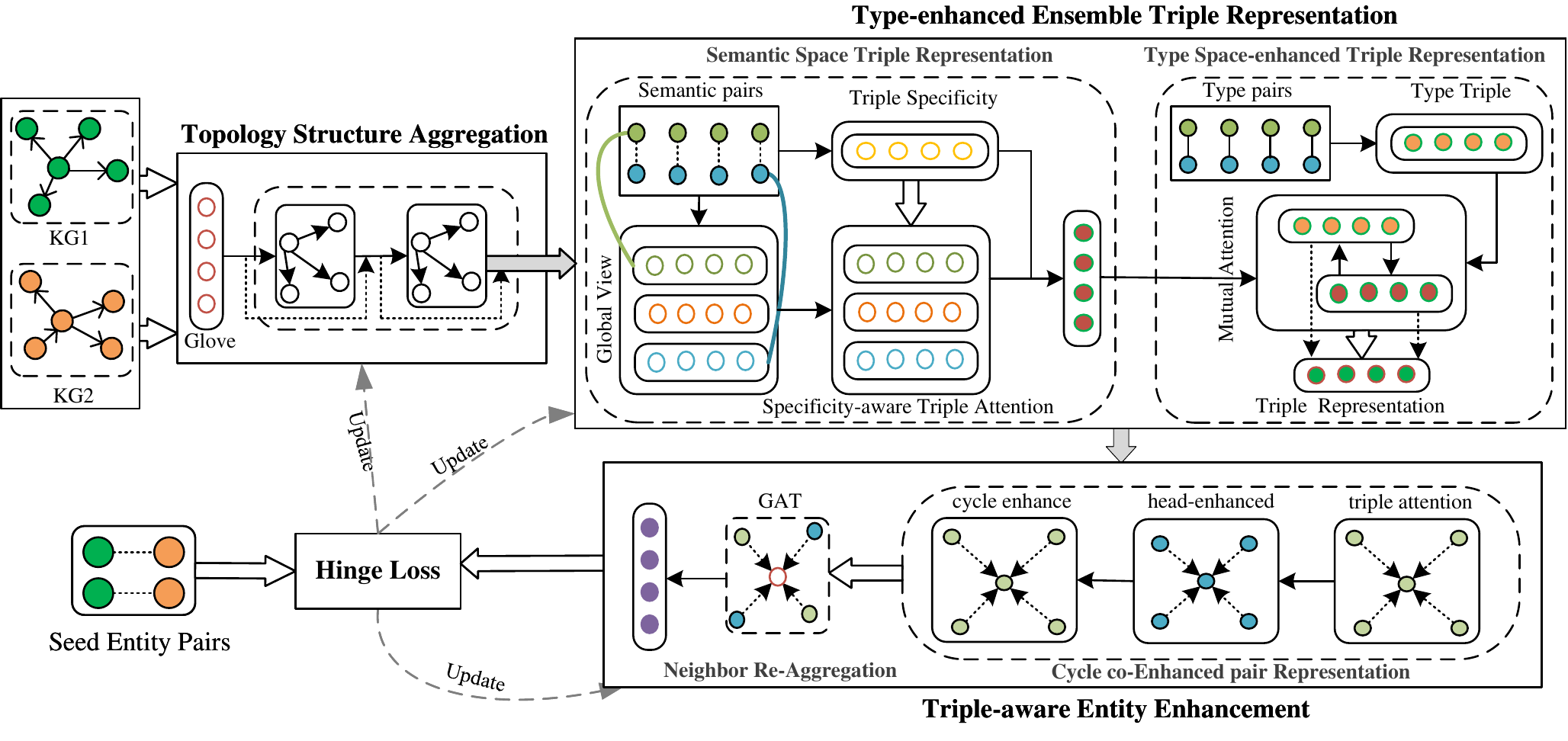}
	\caption{The overall architecture of TTEA framework.}
	\label{FIG:2}
\end{figure*}

We propose a framework TTEA based on type-enhanced ensemble triple representation and triple-aware entity enhancement via triple-aware attention mechanism. The overall architecture of TTEA is illustrated as Figure \ref{FIG:2}, which mainly consists of four parts: Topology Structure Aggregation, Type-enhanced Ensemble Triple Representation, Triple-aware Entity Enhancement and Entity Alignment Strategy. Entity name-based embedding is enhanced via structural information for initialization in the first part, after that the ensemble triple representation with specificity is generated in Type-enhanced Ensemble Triple Representation part. Then, triple-aware representations of head and tail entities are respectively obtained and are circularly reinforced by each other in Triple-aware Entity Enhancement module. Finally, in the Entity Alignment Strategy part, the bi-direction iterative strategy is applied to enlarge seed pairs, meanwhile the entity embedding and the parameters are updated via back-propagation.
\subsection{Topology Structure Aggregation}\label{subsec4}
We firstly expand relation as a combination of original-relation, reverse-relation and self-relation to fully describe topology structure in KGs.
Inspired by RAGA\cite{zhu_raga_2021}, we also use the entity name-based embedding as the initialized representation, following which a two-layer GCNs with Highway Networks\cite{srivastava_highway_2015} are deployed to aggregate topological information while preserving entity primary semantic. The $l$-th Highway-GCN layer is computed as:
\begin{equation}
	\boldsymbol{X}^{l+1}=\operatorname{ReLU}\left(\boldsymbol{\tilde{D}}^{-\frac{1}{2}} \boldsymbol{\tilde{A}} \boldsymbol{\tilde{D}}^{-\frac{1}{2}} \boldsymbol{X}^{l}\right)  \label{eq1}
\end{equation}
\begin{equation}
	T\left(\boldsymbol{X}^{l}\right) = \sigma\left(\boldsymbol{X}^{l} \boldsymbol{W}^{l}+\boldsymbol{b}^{l}\right) \label{eq2}
\end{equation}
\begin{equation}
	\boldsymbol{X}^{l+1}\!=\!T\left(\boldsymbol{X}^{l}\right)\! \cdot \! \boldsymbol{X}^{l+1}\!+\!\left(1\!-\!T\!\left(\!\boldsymbol{X}^{l}\!\right)\!\right)\! \cdot \!\boldsymbol{X}^{l} \label{eq3}
\end{equation}
where $\!\boldsymbol{\tilde{A}}\!\!=\!\boldsymbol{A}\!+\!\boldsymbol{I}$, $\boldsymbol{A}$ is the adjacency matrix of relation-expanded graph, $I$ is the identity matrix, $\boldsymbol{\tilde{D}}$ is the degree matrix of $\boldsymbol{\tilde{A}}$ and $\!\boldsymbol{X}^{l}\!\!\in\! \mathbb{R}^{n\!\times\!d_{e}}$ denotes the input entity embedding in $l$-th hidden layer, $n$ is the number of entities in a KG, $d_{e}$ is the dimension of entity embedding, $\boldsymbol{X}^{l+1}$ is the output of $l$-th layer. $\sigma(\cdot)$ is activation function, $\cdot$ denotes the element-wise multiplication, $\boldsymbol{W}^{l}$ and $\boldsymbol{b}^{l}$ are the weight matrix and bias vector of the input embedding in $l$-th hidden layer.
\subsection{Type-enhanced Ensemble Triple Representation}\label{subsec4}
The ensemble triple representation is generated in this module via mining the internal correlation of triple in semantic space and type space.
\subsubsection*{Ensemble Triple Representation in Semantic Space}\label{subsubsec4}
To describe relational wholeness and multi-relation features, the global relation $\boldsymbol{\bar{r}}\!\in \!\mathbb{R}^{2d_e}$ is computed as the average concatenation of head and tail entities with same relation.
\begin{equation}
	\boldsymbol{\bar{r}}_{r}=\frac{1}{\left|T_{r}\right|} \! \!\sum_{(e_{i}, r, e_{j}) \in T_{r}}\!\left(\boldsymbol{x}_{i} \| \boldsymbol{x}_{j}\right)\! \label{eq4}
\end{equation}
where $\boldsymbol{x}_{i}$ is the embedding of $e_{i}$, $\|$ denotes the concatenation operation, $T_{r}$ is the set of triple with a specific relation $r$ in KG.

Moreover, we utilize triple specificity to alleviate the redundancy and noise during element interaction. The local relation of a specific triple $(e_{i}, r, e_{j})$ is firstly defined as $\boldsymbol{\tilde{r}}_{irj}\!=\!\boldsymbol{x}_{i} \| \boldsymbol{x}_{j}$, based on which the triple specificity is denoted as $\boldsymbol{\tilde{R}}_{irj}\!=\!\boldsymbol{x}_{i} \|(\boldsymbol{\tilde{r}}_{irj}\boldsymbol{W}_{e}^{l}) \| \boldsymbol{x}_{j}$, where $\boldsymbol{W}_{e}^{l}\! \in \!\mathbb{R}^{2d_e\! \times \!d_r}$ is the mapping matrix, $d_r$ is the relation dimension.

Then, we design a head-aware attention, a tail-aware attention and a relation-aware attention to legitimately incorporate global triple features. It is noteworthy that we use the overall relation $\boldsymbol{\tilde{r}}_{r}\!=\!\boldsymbol{W}^{g}_{e}\boldsymbol{\bar{r}}_{r}\! + \!\boldsymbol{W}^{l}_{e}\boldsymbol{\tilde{r}}_{irj}$ for relation-aware triple attention mechanism, where $\boldsymbol{W}^{g}_{e}, \boldsymbol{W}^{l}_{e} \in \mathbb{R}^{2d_e\! \times \!d_r}$ are mapping matrices for global and local relation respectively. Specifically, the head-aware semantic triple representation $\boldsymbol{\bar{x}}_{h r}$ is obtained via head-aware triple attention:
\begin{equation}
	c_{i r}=a^{T}\left(\boldsymbol{x}_{i} \boldsymbol{W}_{h} \|\boldsymbol{\tilde{R}}_{irj} \boldsymbol{W}^{s}_{t}\right) \label{eq5}
\end{equation}
\begin{eqnarray}
	\alpha_{i r}=\frac{\exp \left(\operatorname{LReLU}\left(c_{i r}\right)\right)}{\sum_{\left(e_{i^{\prime}}, r, e_{j^{\prime}} \right) \in T_{r}} \exp \left(\operatorname{LReLU}\left(c_{i^{\prime} r}\right)\right)} \label{eq6}
\end{eqnarray}
\begin{equation}
	\boldsymbol{\bar{x}}_{h r}=\operatorname{LReLU}\left(\sum_{T_r}\left(\alpha_{i r} \boldsymbol{x}_{i} \boldsymbol{W}_{h}\right)\right) \label{eq7}
\end{equation}
where $a\!\in\!\mathbb{R}^{d_r\! \times \!1}$ is a one-dimension vector to map the multi-dimension input into a scalar. $\boldsymbol{W}_{h}\! \in \!\mathbb{R}^{d_e\! \times \!d_r}$ and $\boldsymbol{W}^{s}_{t}\! \in \!\mathbb{R}^{(2d_e\!+\!d_r)\! \times \!d_r}$ are linear transition matrices for head entity and ensemble triple in semantic space. Then the tail-aware semantic triple representation $\boldsymbol{\bar{x}}_{t r}$ and relation-aware semantic triple representation $\boldsymbol{\bar{R}}_{r}$ can be obtained in the same way.

Then, the fused ensemble triple representation in semantic space is computed as combining $\boldsymbol{\bar{x}}_{h r}$, $\boldsymbol{\bar{x}}_{t r}$ and $\boldsymbol{\bar{R}}_{r}$, which is added with the primary specificity as the final triple representation $\boldsymbol{S}_{irj}$ in semantic space for $(e_{i}, r, e_{j}) \in T$ to fully retain the triple semantic specificity:
\begin{equation}
	\boldsymbol{S}_{irj}=\boldsymbol{\bar{x}}_{h r}+\boldsymbol{\bar{x}}_{tr}+\boldsymbol{\bar{R}}_{r}+(\boldsymbol{\tilde{R}}_{irj}\boldsymbol{W}^{s}_{t}) \label{eq13}
\end{equation}
\subsubsection*{Type Space-enhanced Triple Representation}\label{subsubsec4}
In this module, we adopt nonlinear mapping to generate type embedding $\boldsymbol{X}^{t}\! \in \!\mathbb{R}^{n\! \times \!d_t}$ from semantic embedding $\boldsymbol{X}\! \in \!\mathbb{R}^{n\! \times \!d_e}$, where $d_t$ is the type dimension:
\begin{equation}
	\boldsymbol{X}^{t}\!=\!\operatorname{tanh}({\boldsymbol{XW}}\!+\!b) \label{eq14}
\end{equation}

To effectively characterize triple, we regard the elements of type triple as a whole to incorporate type information considering the type inseparability. For a triple $(e_i,r,e_j)$, the global relation representation $\boldsymbol{\bar{r}}^{t}_{r}\! \in \!\mathbb{R}^{2d_t}$ is computed as averaging the concatenation of entity pair with the same relation $r$, which is added to the local relation $\boldsymbol{\tilde{r}}^{t}_{irj}\!=\!\boldsymbol{x_{i}} \| \boldsymbol{x_{j}}$ for generating type triple $\boldsymbol{\tilde{R}}^{t}_{irj}$ as Eq \eqref{eq15}-\eqref{eq17}, where $\boldsymbol{W}^{r}_t \in \mathbb{R}^{2d_t\! \times \!d_r}$ is the transition matrix:
\begin{equation}
	\boldsymbol{\bar{r}}^{t}_{r}=\frac{1}{\left|T_{r}\right|} \sum_{(e_i,r,e_j) \in T_{r}}\left(\boldsymbol{x}^{t}_{i} \| \boldsymbol{x}^{t}_{j}\right) \label{eq15}
\end{equation}
\begin{equation}
	\boldsymbol{r}^{t}_{irj}=\boldsymbol{\bar{r}}^{t}_{r}+\boldsymbol{\tilde{r}}^{t}_{irj} \label{eq16}
\end{equation}
\begin{equation}
	\boldsymbol{\tilde{R}}^{t}_{irj} = \boldsymbol{x}^{t}_{i} \| (\boldsymbol{r}^{t}_{irj}\boldsymbol{W}^{r}_t) \| \boldsymbol{x}^{t}_{j} \label{eq17}
\end{equation}

Then a Semantic-Type mutual attention is designed, in which the enhanced type-space triple representation $\boldsymbol{\bar{T}}_{r}$ can be obtained:
\begin{eqnarray}
	\!\alpha_{t r}\!\!=\!\frac{\exp \left(\!\operatorname{LReLU}\!\!\left(\!a^{T}\!\left(\!\boldsymbol{S}_{irj} \| \boldsymbol{\tilde{R}}^{t}_{irj} \boldsymbol{W}_{t}\!\right)\right)\right)\!}{\!\sum\limits_{T_r}\! \exp \left(\!\operatorname{LReLU}\!\!\left(\!a^{T}\!\left(\!\boldsymbol{S}_{i^{\prime}rj^{\prime}} \| \boldsymbol{\tilde{R}}^{t}_{i^{\prime}rj^{\prime}} \boldsymbol{W}_{t}\!\right)\right)\right)\!} \label{eq18}
\end{eqnarray}
\begin{equation}
	\boldsymbol{\bar{T}}_{r}=\operatorname{ReLU}\left(\sum_{(e_i,r,e_j) \in T_{r}}\left(\alpha_{t r} \boldsymbol{S}_{irj}\right)\right) \label{eq19}
\end{equation}
where $\boldsymbol{W}_{t}\! \in \!\mathbb{R}^{(2d_t\!+\!d_r)\!\times \!d_r}$ are the trainable parameter for type triple. And the enhanced global representation $\boldsymbol{\bar{S}}_{r}$ can be generated in the same way.

Finally, the type-enhanced ensemble triple representation $\boldsymbol{T}_{ijr}\! \in \!\mathbb{R}^{(d_r\!+\!d_t)}$ is obtained while preserving primary type features.
\begin{equation}
\boldsymbol{T}^{\prime}_{irj}\!=\!\boldsymbol{S}_{irj}\!+\!\boldsymbol{\bar{S}}_{r}\!+\!(\boldsymbol{\tilde{R}}^{t}_{irj}\boldsymbol{W}_{t})\!+\!\boldsymbol{\bar{T}}_{r}
\end{equation}
\begin{equation}
	\boldsymbol{T}_{irj} = \boldsymbol{T}^{\prime}_{irj} \| \boldsymbol{\bar{r}}^{t}_{r} \label{eq20}
\end{equation}
\subsection{Triple-aware Entity Enhancement}\label{subsec4}
\subsubsection*{Cycle co-Enhanced Entity Pair Representation}\label{subsubsec4}
An entity may play different roles as head or tail in different triples and the influences of a head entity $e_i$ and a tail entity $e_j$ on triple $(e_i, r, e_j)$ are entirely different. In this module, head entity and tail entity are respectively generated via triple attention and get reinforced circularly.

\begin{eqnarray}
	\alpha_{t h}\!=\! \frac{\!\exp \!\left(\!\operatorname{LReLU}\!\!\left(\!a^{T}\!\left(\!\boldsymbol{T}_{ijr}\boldsymbol{W}^{c}_{h} \| \boldsymbol{x}_{i}\!\right)\right)\right)\!}{\!\sum\limits_{\!\left(\!e_i,r^{\prime},e_j^{\prime}\!\right)\! \in T}\!\!\exp \!\left(\!\operatorname{LReLU}\!\!\left(\!a^{T}\!\left(\!\boldsymbol{T}_{ij^{\prime}r^{\prime}}\boldsymbol{W}^{c}_{h} \| \boldsymbol{x}_{i}\!\right)\right)\right)\!} \label{eq21}
\end{eqnarray}
\begin{equation}
	\boldsymbol{x}_{i}\!=\!\boldsymbol{x}_{i}\!+\!\operatorname{ReLU}\!\left(\!\sum_{\left(e_i,r,e_j\right) \in T}\left(\alpha_{t h} \boldsymbol{T}_{ijr} \boldsymbol{W}^{c}_{h}\right)\!\right)\! \label{eq22}
\end{equation}
\begin{eqnarray}
	\alpha_{t t}\!=\!\frac{\!\exp \!\left(\!\operatorname{LReLU}\!\!\left(\!a^{T}\!\left(\!\boldsymbol{T}_{ijr} \boldsymbol{W}^{c}_{t} \| \boldsymbol{x}_{j}\!\right)\right)\right)\!}{\!\sum\limits_{\!\left(\!e_i^{\prime},r^{\prime},e_j\!\right)\! \in T}\!\! \exp \!\left(\!\operatorname{LReLU}\!\!\left(\!a^{T}\!\left(\!\boldsymbol{T}_{i^{\prime}jr^{\prime}} \boldsymbol{W}^{c}_{t} \| \boldsymbol{x}_{j}\!\right)\right)\right)\!} \label{eq21}
\end{eqnarray}
\begin{equation}
	\boldsymbol{x}_{j}\!=\!\boldsymbol{x}_{j}\!+\!\operatorname{ReLU}\left(\sum_{\left(e_i,r,e_j\right) \in T}\left(\alpha_{t t} \boldsymbol{T}_{ijr} \boldsymbol{W}^{c}_{t}\right)\!\right)\! \label{eq22}
\end{equation}

where $\boldsymbol{W}^{c}_{h}, \boldsymbol{W}^{c}_{t} \in \mathbb{R}^{(d_r\!+\!d_t)\! \times \!d_e}$ are the trainable weight parameters for triples.
\subsubsection*{Neighbor Re-Aggregation}\label{subsubsec4}
In the last part of TTEA, we apply a GAT layer to re-aggregate neighbor information with modeled representation and the final entity representation $\boldsymbol{x}_{i}^{f}$ is generated for EA.

\begin{equation}
	\alpha_{i j}=\frac{\exp \left(\operatorname{LReLU }\left(a^{T}\left(\boldsymbol{x}_{i} \| \boldsymbol{x}_{j}\right)\right)\right)}{\sum\limits_{e_k \in N_{i}} \exp \left(\operatorname{LReLU}\left(a^{T}\left(\boldsymbol{x}_{i} \| \boldsymbol{x}_{k} \right)\right)\right)} \label{eq23}
\end{equation}
\begin{equation}
	\boldsymbol{x}_{i}^{f}=\boldsymbol{x}_{i}  \| \left( \operatorname{ReLU}\left(\sum_{e_j \in N_{i}}\left(\alpha_{i j} \boldsymbol{x}_{j}\right)\right)\right) \label{eq24}
\end{equation}

where $N_{i}$ is the set of neighbor entities of $e_i$.
\subsection{Entity Alignment Strategy}\label{subsec4}
Manhattan distance is adopt to measure the similarity of entities, based on which the margin-based loss $L$ is defined as Eq \eqref{eq26}. Moreover, we deploy a bi-direction iterative strategy following MRAGA\cite{mao_mraea_2020} to expand training seed pairs based on negative-sample method.

\begin{equation}
	{dis}\left(e_{i}, e_{j}\right)=\left\|\boldsymbol{x}_{i}^{f}-\boldsymbol{x}_{j}^{f}\right\|_{1} \label{eq25}
\end{equation}
\begin{eqnarray}
L\!= \!\sum_{\!\left(\!e_{i}, e_{j}\!\right)\! \in S}\max\!\left(\!{dis}\!\left(\!e_{i}, e_{j}\!\right)\!\!-\!{dis}\!\left(\!e_{i}^{\prime}, e_{j}^{\prime}\!\right)\!\!+\!\lambda, 0\!\right)\! \label{eq26}
\end{eqnarray}

where $(e_{i}, e_{j})$ is a pre-aligned entity pair in training set $S$, $(e_i^{\prime}, e_j^{\prime})$ is the negative sample generated by randomly replacing $e_{i}$ or $e_{j}$ with their $k$-nearest neighbors, $\lambda$ is the margin hyper-parameter.

\section{Experimental Setup}\label{sec5}
\subsection{DataSets}\label{subsec5}
In order to make the reliable and fair comparison with previous methods, we evaluate TTEA on three real-world multi-lingual datasets from simplified DBP15K described in Table \ref{lab1}, which is constructed by removing lots of unrelated entities and relations from initial DBP15K and is adopted by almost all related works.
\begin{table}[h]
	\caption{Statistical data of Simplified DBP15K.} \label{lab1}
	\resizebox{\linewidth}{!}{
	\begin{tabular}{cccccc}
		\toprule%
		{DBP15K} &{} & {Entities} & {Relations} & {Rel Triples} & {Links}\\
		\midrule
		\multirow{2}{*}{ZH-EN}
		&{ZH} & {19388} & {1700} & {70414}& \multirow{2}{*}{15000}\\
		&{EN} & {19572} & {1322} & {95142} &\\
		\midrule
		\multirow{2}{*}{JA-EN} 
		&{JA} & {19814} & {1298} & {77214} &\multirow{2}{*}{15000}\\
		&{EN} & {19780} & {1152} & {93484} &\\
		\midrule
		\multirow{2}{*}{FR-EN} 
		&{FR} & {19661} & {902} & {105998} &\multirow{2}{*}{15000}\\
		&{EN} & {19993} & {1207} & {115722} &\\
		\bottomrule 
	\end{tabular}
}
\end{table}
\subsection{Baselines}\label{subsec5}
To comprehensively evaluate our approach, we compare TTEA with Trans-based, GNNs-based and Semi-supervised entity alignment methods:

\textbf{-- Trans-based methods:} MTransE\citep{chen_multilingual_2017}, JAPE\citep{JAPE}, BootEA\citep{sun_bootstrapping_2018}, TransEdge\citep{sun_transedge_2019}, RpAlign\citep{huang_cross-knowledge-graph_2022}.

\textbf{-- GNNs-based methods:}
\textbf{\emph{(1) GCN-based methods:}} GCN-Align\citep{wang_cross-lingual_2018}, HMAN\citep{yang2019aligning}, HGCN\citep{wu2019jointly}, MCEA\citep{qi_multiscale_2022-5}.
\textbf{\emph{(2) GAT-based methods:}} NAEA\citep{zhu_neighborhood-aware_2019}, RDGCN\citep{wu_relation-aware_2019}, NMN\citep{wu_neighborhood_2020}, KAGNN\citep{huang_multi-view_2022}, MRGA\citep{ding_multi_2021}, SHEA\citep{yan_soft-self_2021}, RAGA-l\citep{zhu_raga_2021}.

\textbf{-- Semi-supervised methods:} 
MRAEA\citep{mao_mraea_2020},
RREA-semi\citep{mao_relational_2020}, RAGA-semi\citep{zhu_raga_2021}, MCEA-semi\citep{qi_multiscale_2022-5}.

It should be noted that methods requiring additional information such as RAEA\cite{zhu_cross-lingual_2022} and RNM\cite{zhu_relation-aware_2020} are not considered as baselines for fairness.
\subsection{Model variants}\label{subsec5}
To make valid evaluation on different components in our framework, we implement three variants of TTEA to verify their effectiveness:


(1) wo-E: a simplified TTEA version without Ensemble Triple Attention.

(2) wo-T: a simplified TTEA version without Type Space-Enhanced module.

(3) wo-C: a simplified TTEA version without Cycle co-Enhanced module.

\subsection{Implementation Details}\label{subsec5}
We use Glove\cite{pennington_glove_2014} to generate the initial entity embedding. For a fair comparison with the baselines, we use a 30\% proportion of alignment seeds for training and the rest for testing. The depth $l$ of Highway-GCNs is 2, both the relation dimension $d_r$ and type dimension $d_t$ are 100. And the number of epochs $p$ for updating negative samples is 5, the number of nearest negative samples $k$ is 5. The margin hyper-parameter $\lambda$ is 3.0.
\subsection{Metrics}\label{subsec5}

\begin{table*}[!h]
	\caption{Overall performance of entity alignment.\label{tab2}}
	\centering
	\resizebox{\linewidth}{!}{
	\begin{tabular*}{500pt}{@{\extracolsep{\fill}}lccccccccc@{\extracolsep{\fill}}}
		\toprule%
		& \multicolumn{3}{@{}c@{}}{\textbf{ZH-EN}} & \multicolumn{3}{@{}c@{}}{\textbf{JA-EN}} & \multicolumn{3}{@{}c@{}}{\textbf{FR-EN}} \\\cmidrule{2-4}\cmidrule{5-7} \cmidrule{8-10}%
		\textbf{Methods}& H@1 &H@10 &MRR& H@1 &H@10 &MRR &H@1& H@10 &MRR\\
		\midrule
		MTransE(2017) &30.8 &61.4 &0.364 &27.8 &57.4 &0.349& 24.4 &55.5&0.335\\
		JAPE(2017) &41.2 &74.4& 0.490& 36.2 &68.5& 0.476 &32.4& 66.7 &0.430\\
		BootEA(2018) &62.9& 84.7 &0.703 &62.2& 85.4& 0.701& 65.3 &87.4 &0.731\\
		TransEdge(2019) &73.5 &91.9 &0.801& 71.9 &93.2 &0.795 &71.0 &94.1& 0.796\\
		RpAlign(2022) &74.8 &88.8 &0.794 &72.9& 89.0& 0.872& 75.2 &89.9& 0.801\\
		\midrule
		GCN-Align(2018) &41.2& 74.4& 0.549 &39.9& 74.4 &0.546& 37.3& 74.5& 0.532\\
		HMAN(2019) &56.1 &85.9& 0.67 &55.7& 86.0 &0.67 &55.0& 87.6 &0.66\\
		HGCN(2019) &72.0 &85.7& 0.768 &76.6& 89.7 &0.813 &89.2& 96.1 &0.917\\
		MCEA(2022) &72.4 &93.4 &0.800 &71.9& 94.0& 0.800& 73.9 &95.3& 0.820\\
		\hdashline[0.5pt/2pt]\noalign{\vskip 0.5mm}
		NAEA(2019) &65.0 &86.7 &0.720& 64.1 &87.3 &0.718 &67.3 &89.4& 0.752\\
		RDGCN(2019) &70.8& 84.6 &0.751 &76.7& 89.5& 0.812& 88.6& 95.7& 0.908\\
		NMN(2020) &73.3& 86.9 &0.781 &78.5& 91.2 &0.827 &90.2 &96.7 &0.924\\
		KAGNN(2022) &73.6 &87.3 &0.786 &79.4& 91.1& 0.837& 92.0 &97.6& 0.941\\
		MRGA(2021) &75.5& 90.5 &0.783 &73.4& 90.3 &0.771 &75.7 &91.7 &0.791\\
		SHEA(2021) &76.3 &91.4 &0.835 &82.1& 93.8& 0.860& 90.5 &97.0& 0.902\\
		RAGA-l(2021) &79.8 &93.3 &0.847 &82.9& 95.0& 0.875& 91.4 &98.2& 0.940\\
		\hdashline[5pt/5pt]\noalign{\vskip 0.5mm}
		TTEA-base(wo-E) &78.9 &93.4 &0.842 &82.0& 95.0& 0.868& 91.9 &98.5& 0.944\\   
		TTEA-base(wo-T) &78.7 &93.4 &0.841 &81.4& 95.1& 0.864& 91.4 &98.4& 0.940\\   
		TTEA-base(wo-C) &79.9 &93.5 &0.849 &82.9& 95.1& 0.875& 92.2 &98.3& 0.946\\   
		\textbf{TTEA-base(ours)} &\textbf{80.2} &\textbf{93.8} &\textbf{0.852} &\textbf{83.1}&\textbf{95.4}&\textbf{0.876}&\textbf{92.4}&\textbf{98.6}&\textbf{0.947}\\
		\midrule
		MRAEA(2020) &75.2& 92.3 &0.824 &75.3 &93.3 &0.825 &78.1& 94.7 &0.843\\
		RREA-semi(2020) &80.1&94.8 &0.857 &80.2 &95.2& 0.858&82.7 &96.6& 0.881\\
		MCEA-semi(2022) &81.4 &95.6 &0.867 &80.7& 95.7& 0.864& 84.1 &97.0& 0.891\\
		RAGA-semi(2021) &85.7 &96.0 &0.896 &88.9& 97.1& 0.920& 94.0 &98.8& 0.958\\
		\textbf{TTEA-semi(ours)} &\textbf{86.3} &\textbf{96.2} &\textbf{0.901} &\textbf{89.2}&\textbf{97.6}&\textbf{0.924}&\textbf{94.7}&\textbf{99.0}&\textbf{0.964}\\
		\bottomrule
	\end{tabular*}
}
\end{table*}
By convention, we report the Hits@1, Hits@10 and {MRR} results to evaluate the EA performance. Hits@k measures the percentage of correct alignment ranked at top k, and MRR is the average of the reciprocal ranks of results. Higher Hits@k and MRR scores indicate better performance.
\section{Results and Analysis}\label{sec6}
\subsection{Overall Performance}\label{subsec6}
The results of baselines on three datasets are listed in Table \ref{tab2}, which are either implemented with the source codes or provided by original papers. The solid lines separate Trans-based methods, GNNs-based methods and Semi supervised-based methods and dot line makes a distinction between GCNs-based methods and GATs-based methods and TTEA variants are under the dashed line.

For Trans-based methods, TransEdge and RpAlign outperform MTransE, JAPE, BootEA and NAEA with their unique representation for triple elements. In detail, RpAlign achieves better Hits@1 for its relation prediction and self-training mechanism, while TransEdge gets more excellent Hits@10 and MRR on ZH\_EN and JA\_EN via contextualizing relation representation in terms of specific head-tail entity pair. For GCNs-based methods, GCN-Align gets the worst results as shallow utilization of relation triple while MCEA outperforms others for extending the convolution region of long-tail entities. Furthermore, NAEA, RDGCN, NMN, KAGNN, MRGA, SHEA, RAGA-l all adopt GATs to obtain fine-grained representation, which get excellent performance without doubt. Among them, RAGA-l achieves the best results, which generate relation from entity via attention mechanism and then aggregate relation to entity. Compared with baselines, our TTEA performs best in all evaluation metrics on three datasets with the consideration of triple specificity and the role diversity.
\subsection{Ablation Analysis}\label{subsec6ablation}
\subsubsection*{Effect of TTEA Components}\label{subsubsec6ablation}
The results of TTEA-base(wo-E), TTEA-base(wo-T) and TTEA-base(wo-C) in Table \ref{tab2} show that while ensemble triple attention, Type Space-enhanced module and Cycle co-Enhanced module in TTEA all make a improvement, Type Space-Enhanced module has a more significant effect. Moreover, three modes of Cycle co-Enhanced module with different cycle orders are compared in Table \ref{cycle} to explore appropriate cycle form.

\textit{\textbf{mode1}}: the mode of co-enhanced process with the head-tail order.

\textit{\textbf{mode2}}: the mode of cycle co-enhanced process with the head-tail-head order.

\textit{\textbf{mode3}}: the mode of cycle co-enhanced process with the head-tail-head-tail order.

We can see from Table \ref{cycle} that the mode2 adopt in TTEA is more effective on ZH\_EN and FR\_EN for reasonable cycle process while the mode3 gets the almost same performance on JA\_EN.
\begin{table}[!h]
	\caption{Comparison of different modes of Cycle Co-enhanced module.\label{cycle}}
	\resizebox{\linewidth}{!}{
		\begin{tabular}{@{\extracolsep{\fill}}lccccccccc@{\extracolsep{\fill}}}
			\toprule%
			& \multicolumn{3}{@{}c@{}}{\textbf{ZH-EN}} & \multicolumn{3}{@{}c@{}}{\textbf{JA-EN}} & \multicolumn{3}{@{}c@{}}{\textbf{FR-EN}} \\\cmidrule{2-4}\cmidrule{5-7} \cmidrule{8-10}%
			\textbf{Modes}& H@1 &H@10 &MRR& H@1 &H@10 &MRR &H@1& H@10 &MRR\\
			\midrule
			\textit{mode1} &79.8 &93.5 &0.849 &\textbf{83.1} &95.0 &0.875 &92.2 &98.4 &0.946\\
			\textit{mode2} &\textbf{80.2} &\textbf{93.8} &\textbf{0.852} &\textbf{83.1}&\textbf{95.4}&0.876&\textbf{92.4}&\textbf{98.6}&\textbf{0.947}\\
			\textit{mode3} &79.9 &93.6 &0.849 &\textbf{83.1} &\textbf{95.4} &\textbf{0.877} &92.3 &98.5 &\textbf{0.947}\\
			\bottomrule
		\end{tabular}
	}
\end{table}
\begin{table}[!h]
	\caption{Comparison of different depths of Highway-GCNs.\label{gcn}}
	\resizebox{\linewidth}{!}{
		\begin{tabular}{@{\extracolsep{\fill}}lccccccccc@{\extracolsep{\fill}}}
			\toprule%
			& \multicolumn{3}{@{}c@{}}{\textbf{ZH-EN}} & \multicolumn{3}{@{}c@{}}{\textbf{JA-EN}} & \multicolumn{3}{@{}c@{}}{\textbf{FR-EN}} \\\cmidrule{2-4}\cmidrule{5-7} \cmidrule{8-10}%
			\textbf{Depths}& H@1 &H@10 &MRR& H@1 &H@10 &MRR &H@1& H@10 &MRR\\
			\midrule
			\textit{l}=1 &79.4 &92.4 &0.841 &\textbf{83.4} &94.5 &0.875 &\textbf{92.9} &98.4 &\textbf{0.950}\\
			\textit{l}=2 &\textbf{80.2} &\textbf{93.8} &\textbf{0.852} &83.1 &\textbf{95.4} &\textbf{0.876} &92.4 &\textbf{98.6} &0.947\\
			\textit{l}=3 &77.0 &\textbf{93.8} &0.832 &79.4 &94.8 &0.851 &88.6 &97.7 &0.921\\
			\bottomrule
		\end{tabular}
	}
\end{table}
\begin{figure*}[!h]
	\centering
	\subfigure[ZH\_EN]{
		\begin{minipage}{0.31\textwidth}
			\centering
			\includegraphics[width=\textwidth]{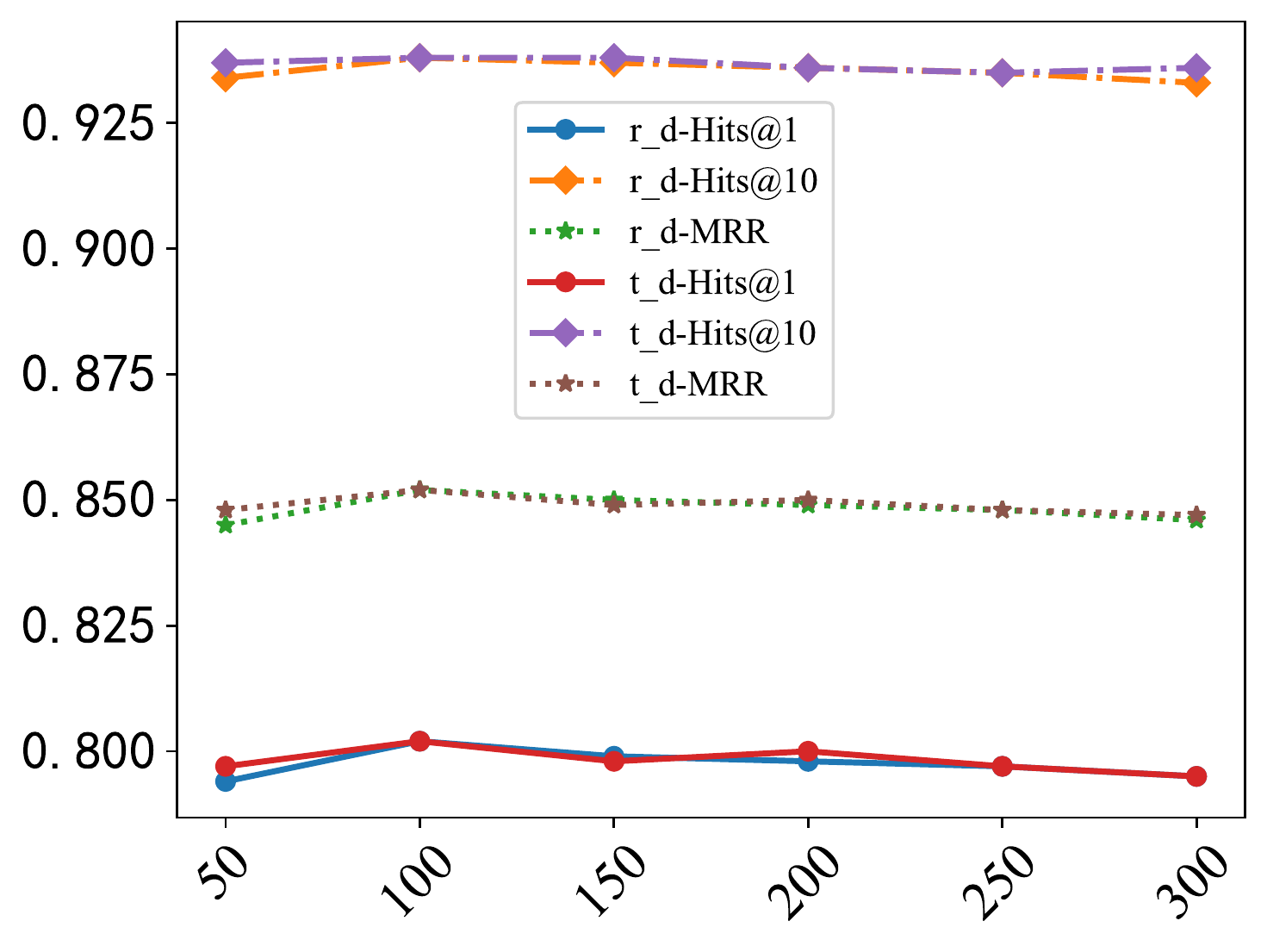}
		\end{minipage}
	}
	\subfigure[JA\_EN]{
		\begin{minipage}{0.31\textwidth}
			\centering
			\includegraphics[width=\textwidth]{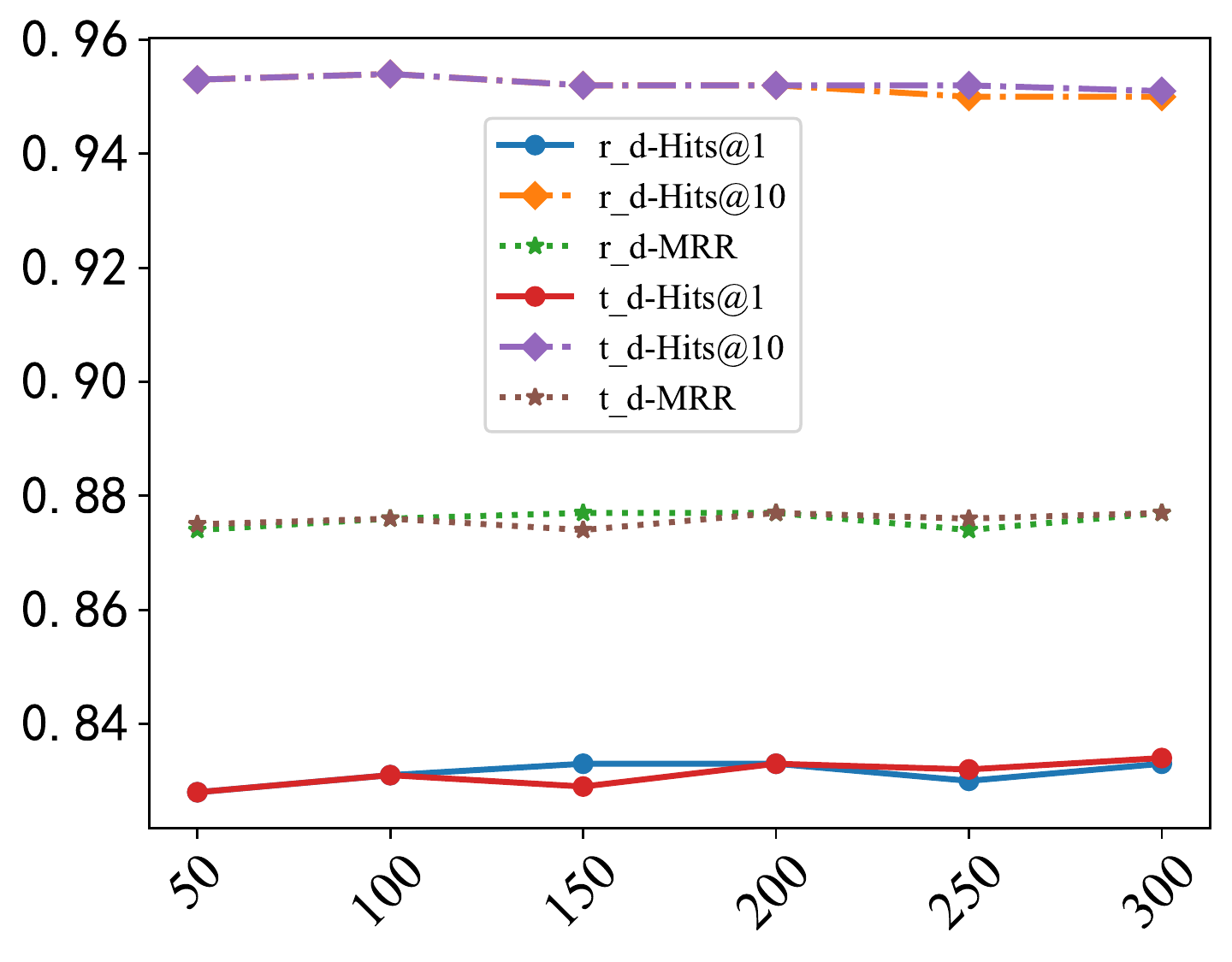}
		\end{minipage}
	}
	\subfigure[FR\_EN]{
		\begin{minipage}{0.31\textwidth}
			\centering
			\includegraphics[width=\textwidth]{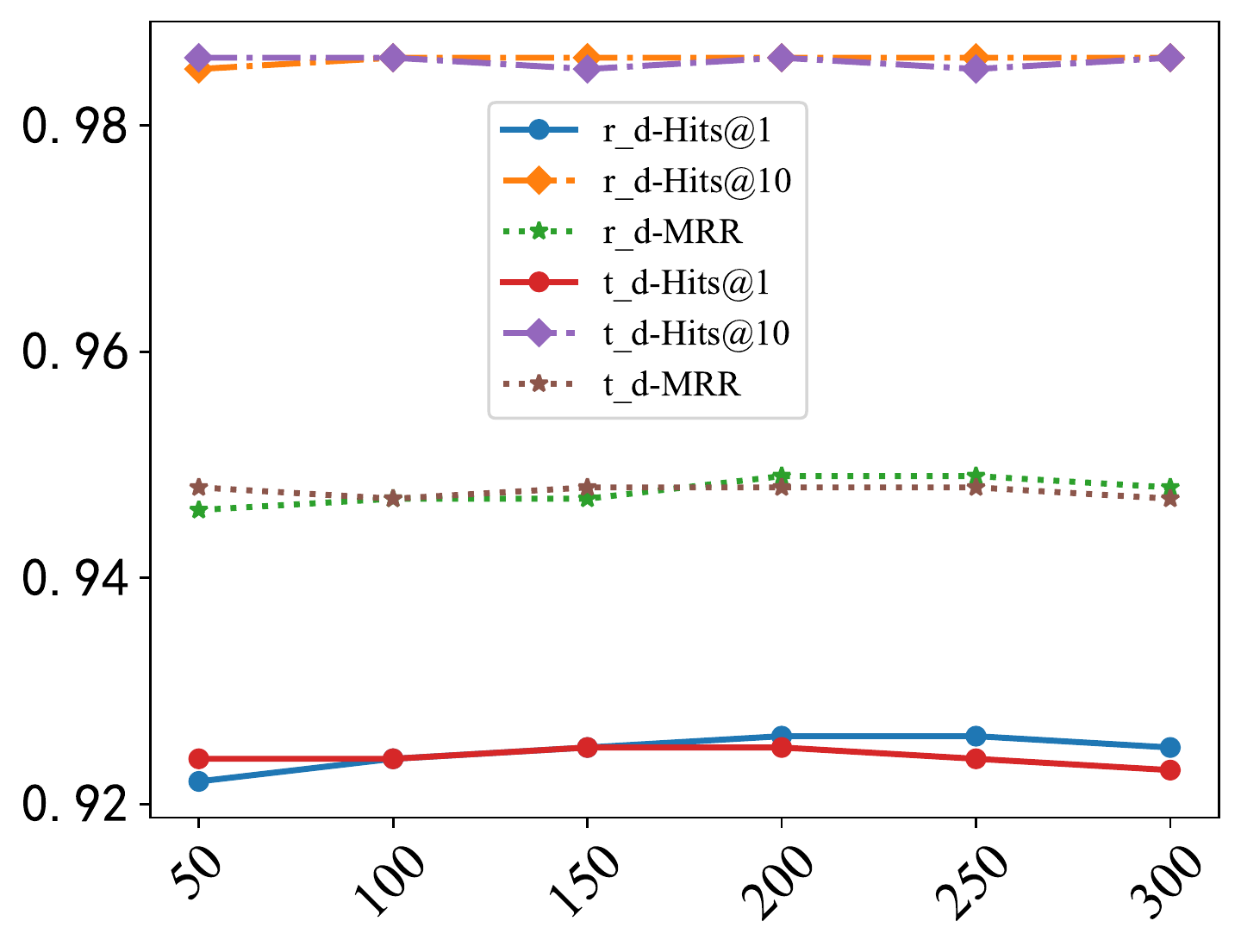}
		\end{minipage}
	}
	\caption{EA performance with different relation and type dimensions.}
	\label{FIG:dimension}
\end{figure*}
\subsubsection*{Impact of GCN depth}\label{subsubsec6ablation}
To explore the impact of different Highway-GCNs depth $l$, we compare TTEA variants with different depths with $l\!=\!1$, $l\!=\!2$ and $l\!=\!3$ in Table \ref{gcn}. The results show that the TTEA variant with a two-layer Highway-GCNs obtains the greatest superiority on ZH\_EN and JA\_EN for their complex structure, while a one-layered Highway-GCNs get the best performance on FR\_EN for its entity semantic reliance.
\subsubsection*{Impact of Relation and Type Dimensions}\label{subsubsec6ablation}
There are two dimensional hyper-parameters: relation dimension $d_r$ and type dimension $d_t$ in TTEA. We respectively evaluate TTEA on six different relation and type dimensions as 50, 100, 150, 200, 250 and 300 to explore dimensional impacts. The results in Figure \ref{FIG:dimension} show that different relation and type dimensions have approximate performance, which indicate that dimensions have little influence on TTEA. Especially, the best results can be obtained on ZH\_EN when $d_r=d_t=100$.

\begin{figure*}[!h]
	\centering
	\subfigure[ZH\_EN]{
		\begin{minipage}{0.3\textwidth}
			\centering
			\includegraphics[width=\textwidth]{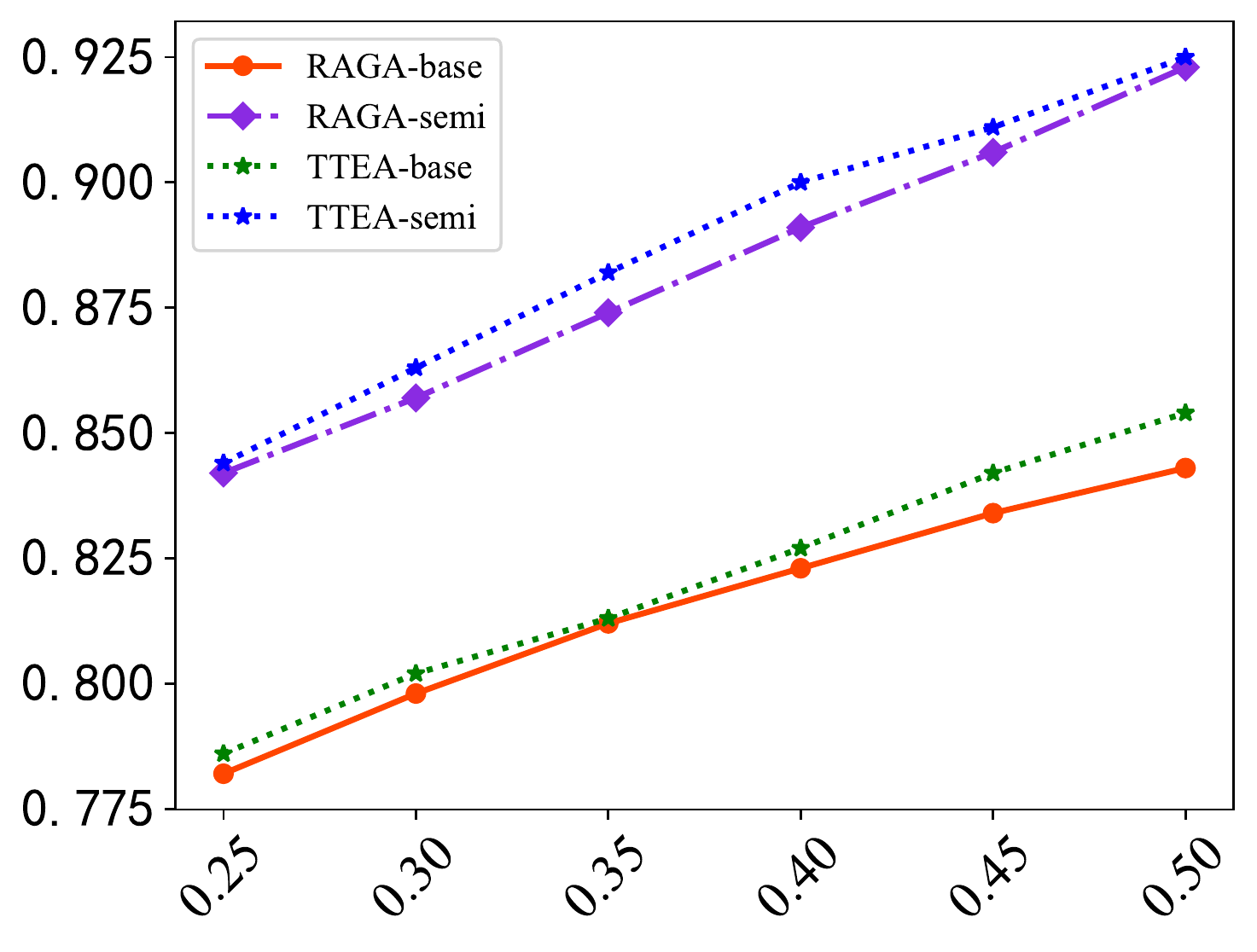}
		\end{minipage}
	}
	\subfigure[JA\_EN]{
		\begin{minipage}{0.3\textwidth}
			\centering
			\includegraphics[width=\textwidth]{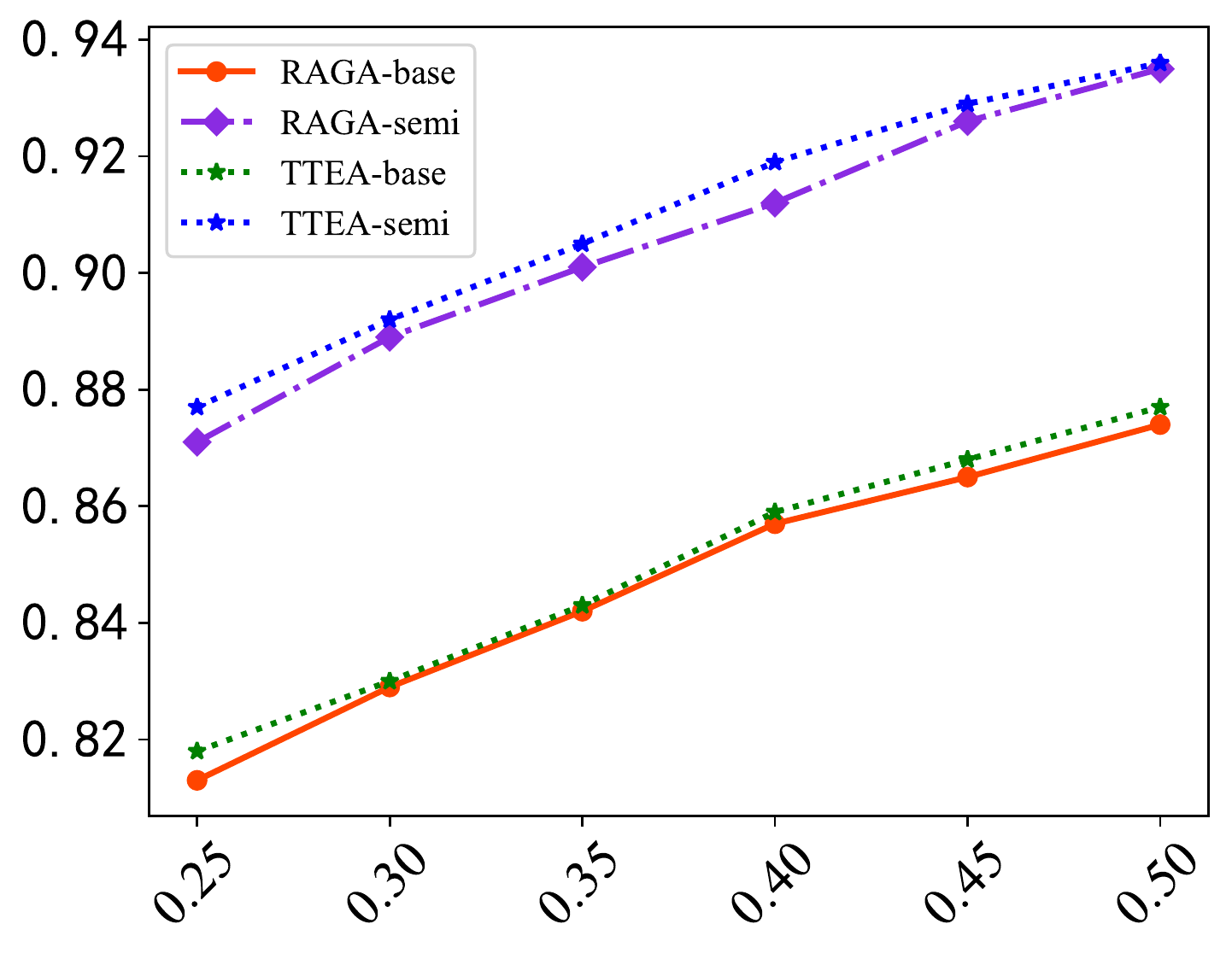}
		\end{minipage}
	}
	\subfigure[FR\_EN]{
		\begin{minipage}{0.3\textwidth}
			\centering
			\includegraphics[width=\textwidth]{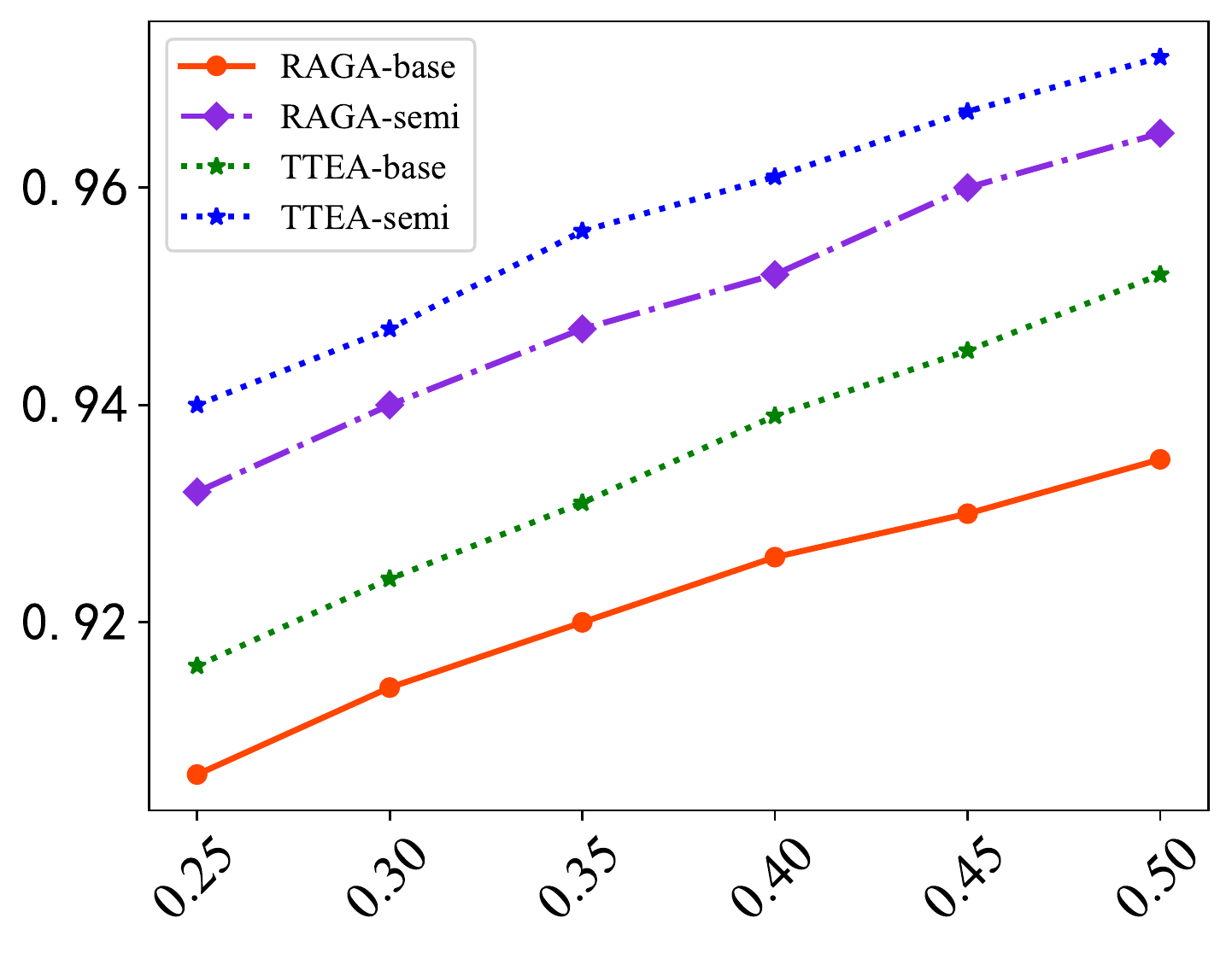}
		\end{minipage}
	}
	\caption{Hits@1 with different training seed pairs.}
	\label{FIG:train1}
\end{figure*}
\begin{figure*}[!h]
	\centering
	\subfigure[ZH\_EN]{
		\begin{minipage}{0.3\textwidth}
			\centering
			\includegraphics[width=\textwidth]{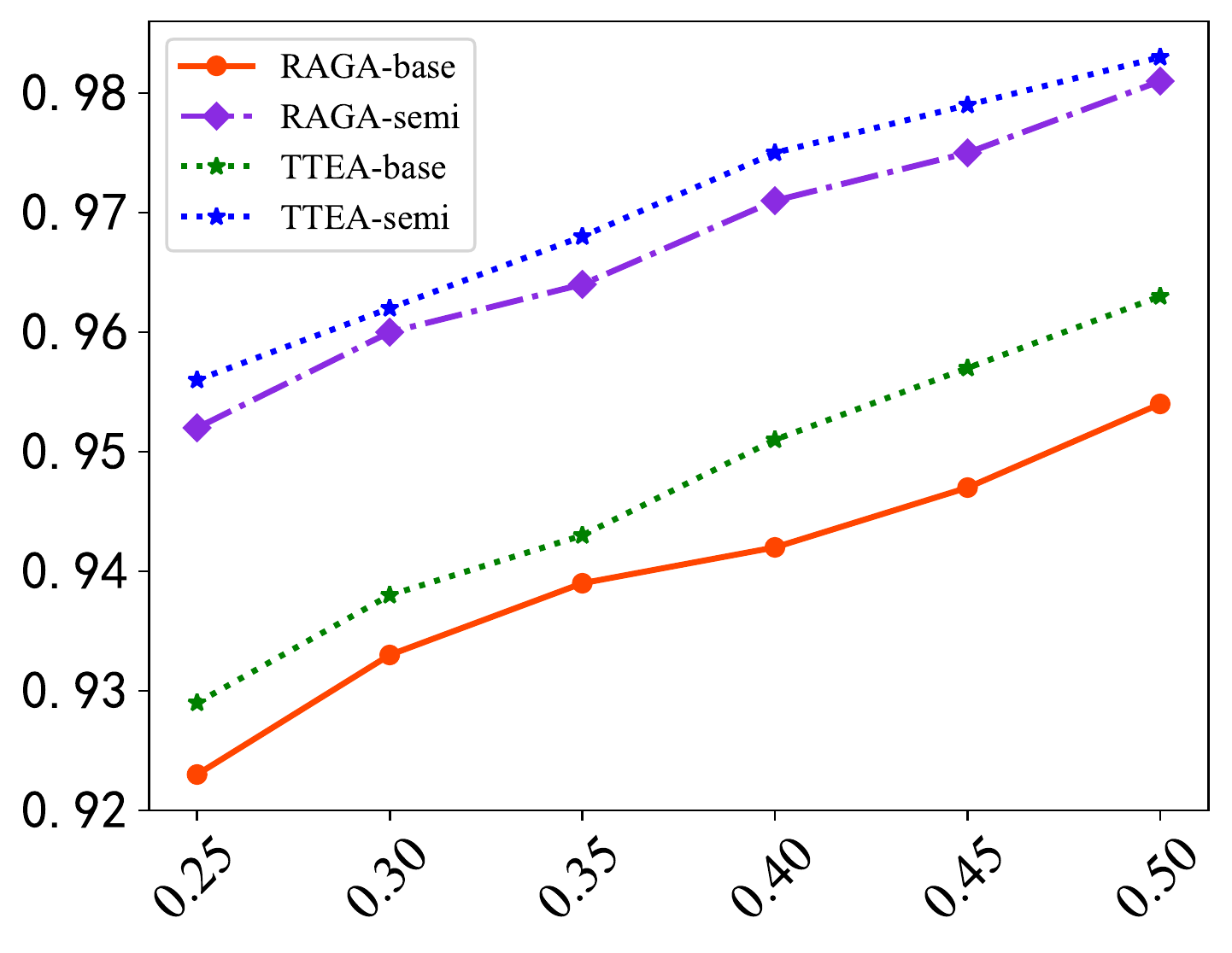}
		\end{minipage}
	}
	\subfigure[JA\_EN]{
		\begin{minipage}{0.3\textwidth}
			\centering
			\includegraphics[width=\textwidth]{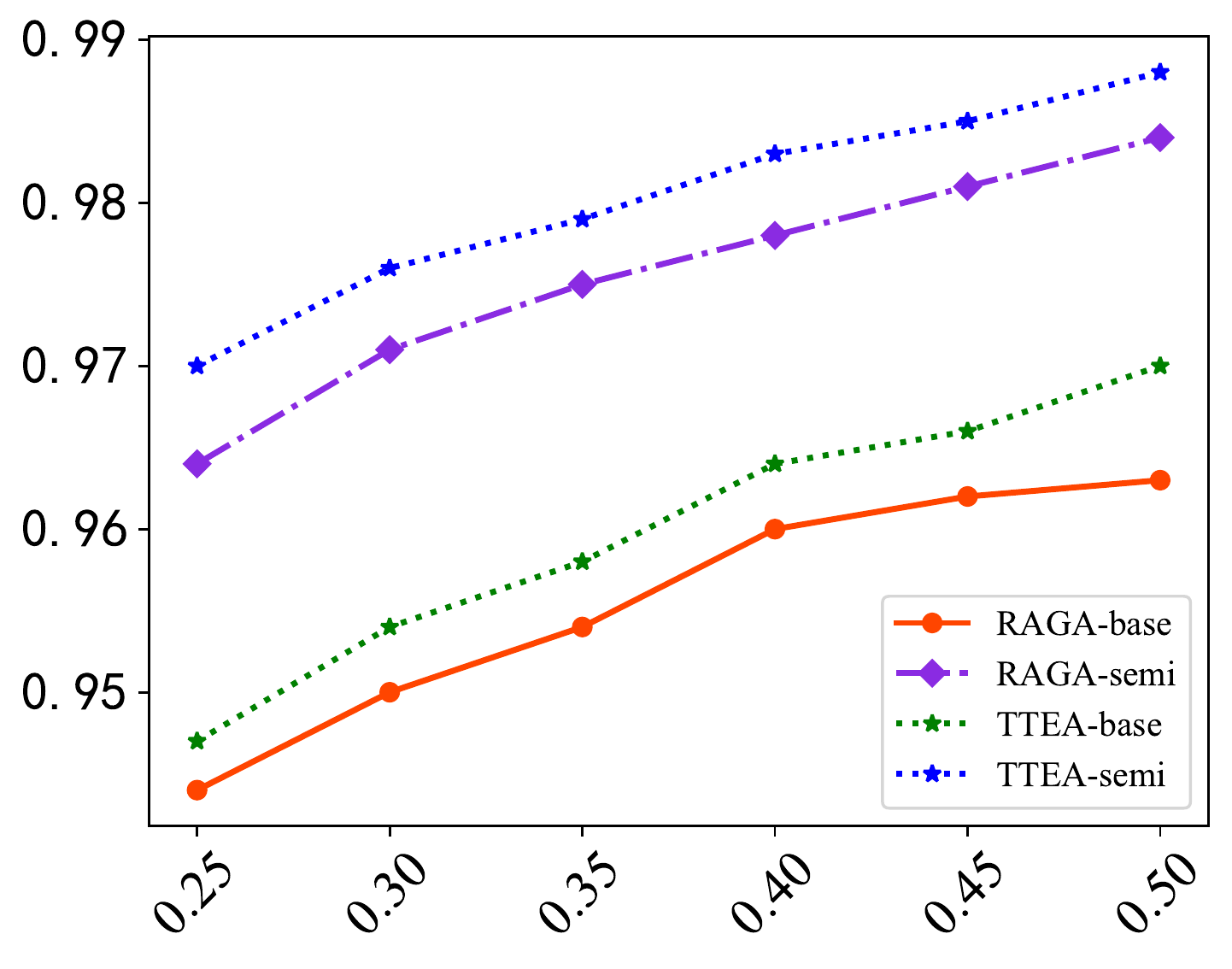}
		\end{minipage}
	}
	\subfigure[FR\_EN]{
		\begin{minipage}{0.3\textwidth}
			\centering
			\includegraphics[width=\textwidth]{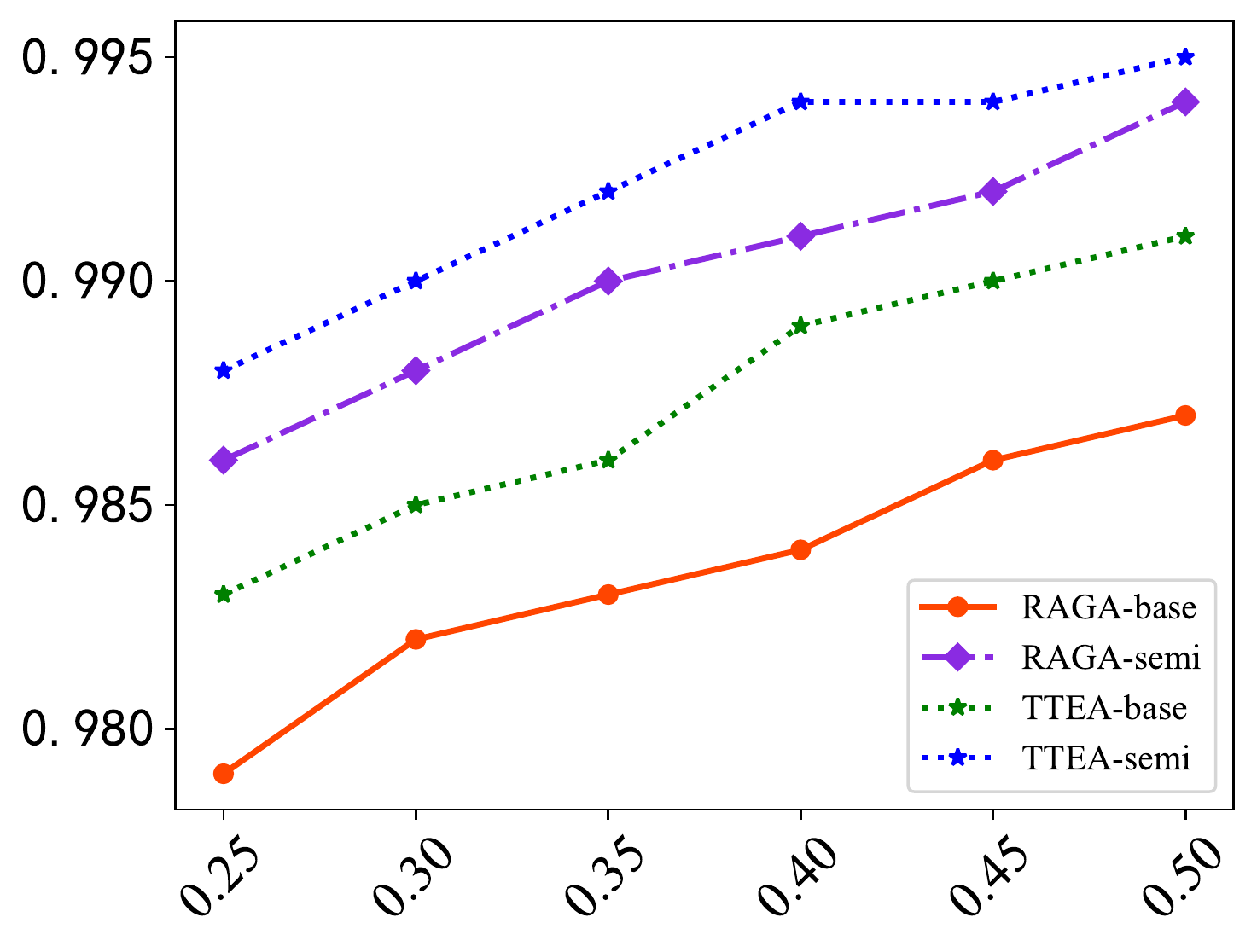}
		\end{minipage}
	}
	\caption{Hits@10 with different training seed pairs.}
	\label{FIG:train10}
\end{figure*}
\begin{figure*}[!h]
	\centering
	\subfigure[ZH\_EN]{
		\begin{minipage}{0.3\textwidth}
			\centering
			\includegraphics[width=\textwidth]{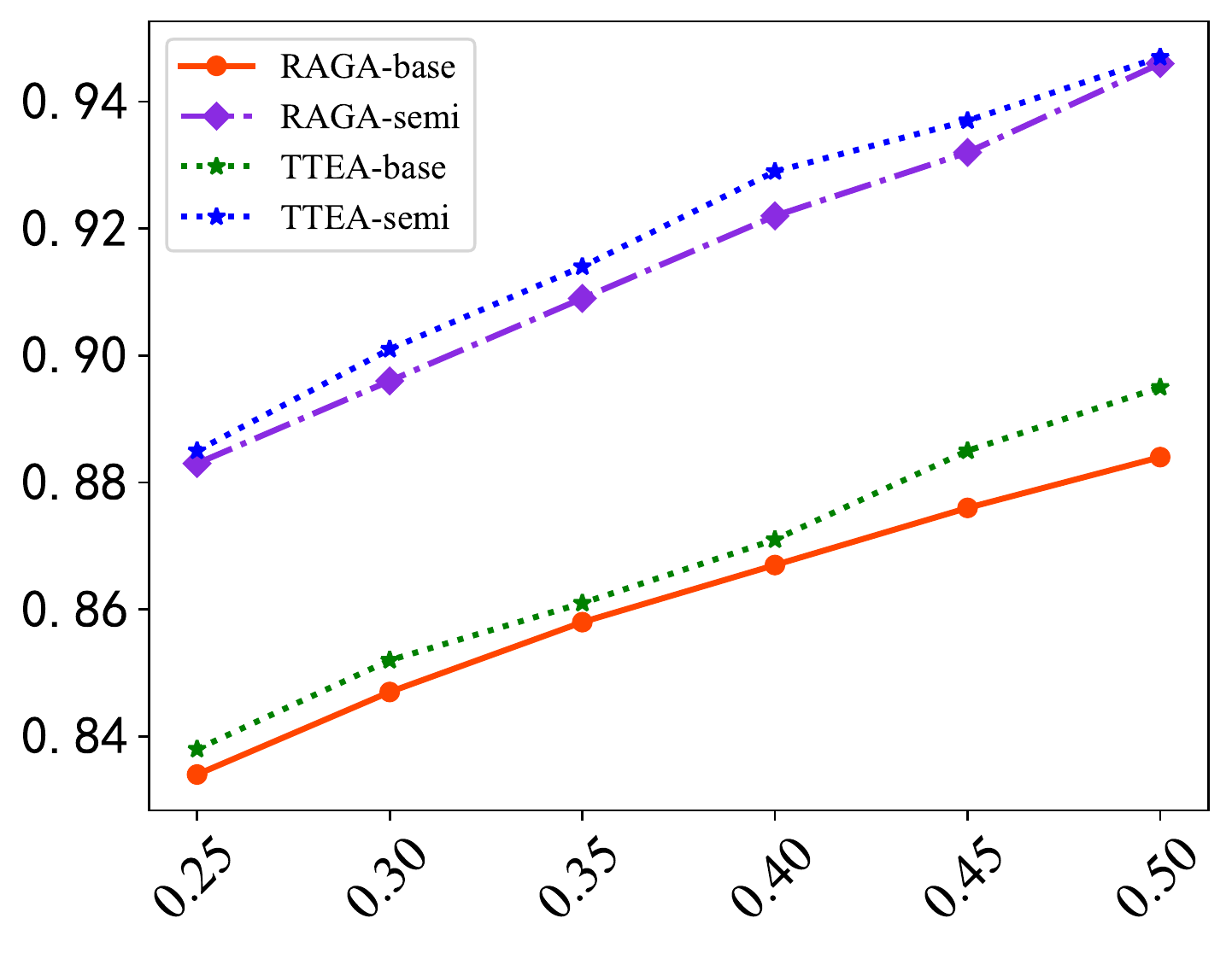}
		\end{minipage}
	}
	\subfigure[JA\_EN]{
		\begin{minipage}{0.3\textwidth}
			\centering
			\includegraphics[width=\textwidth]{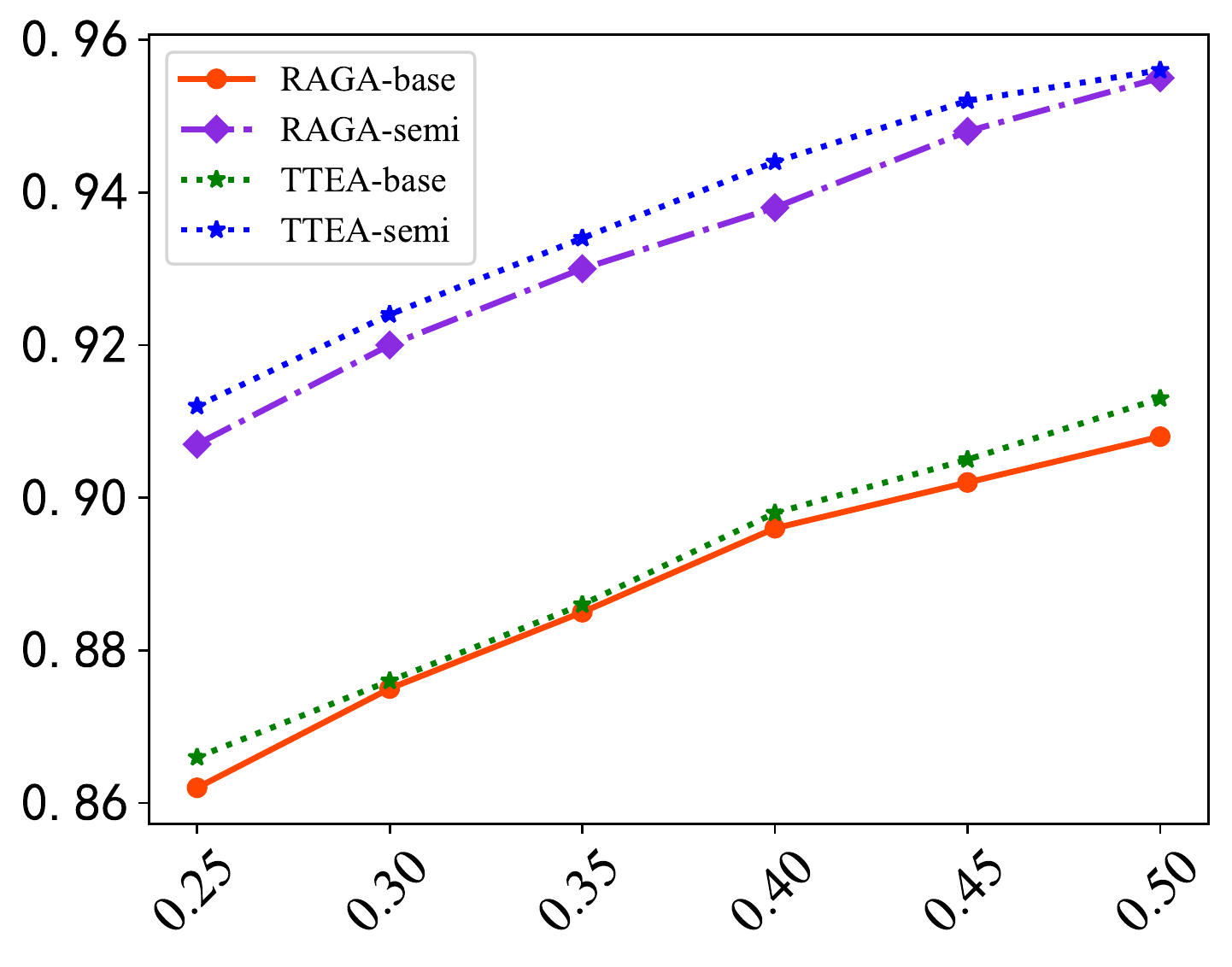}
		\end{minipage}
	}
	\subfigure[FR\_EN]{
		\begin{minipage}{0.3\textwidth}
			\centering
			\includegraphics[width=\textwidth]{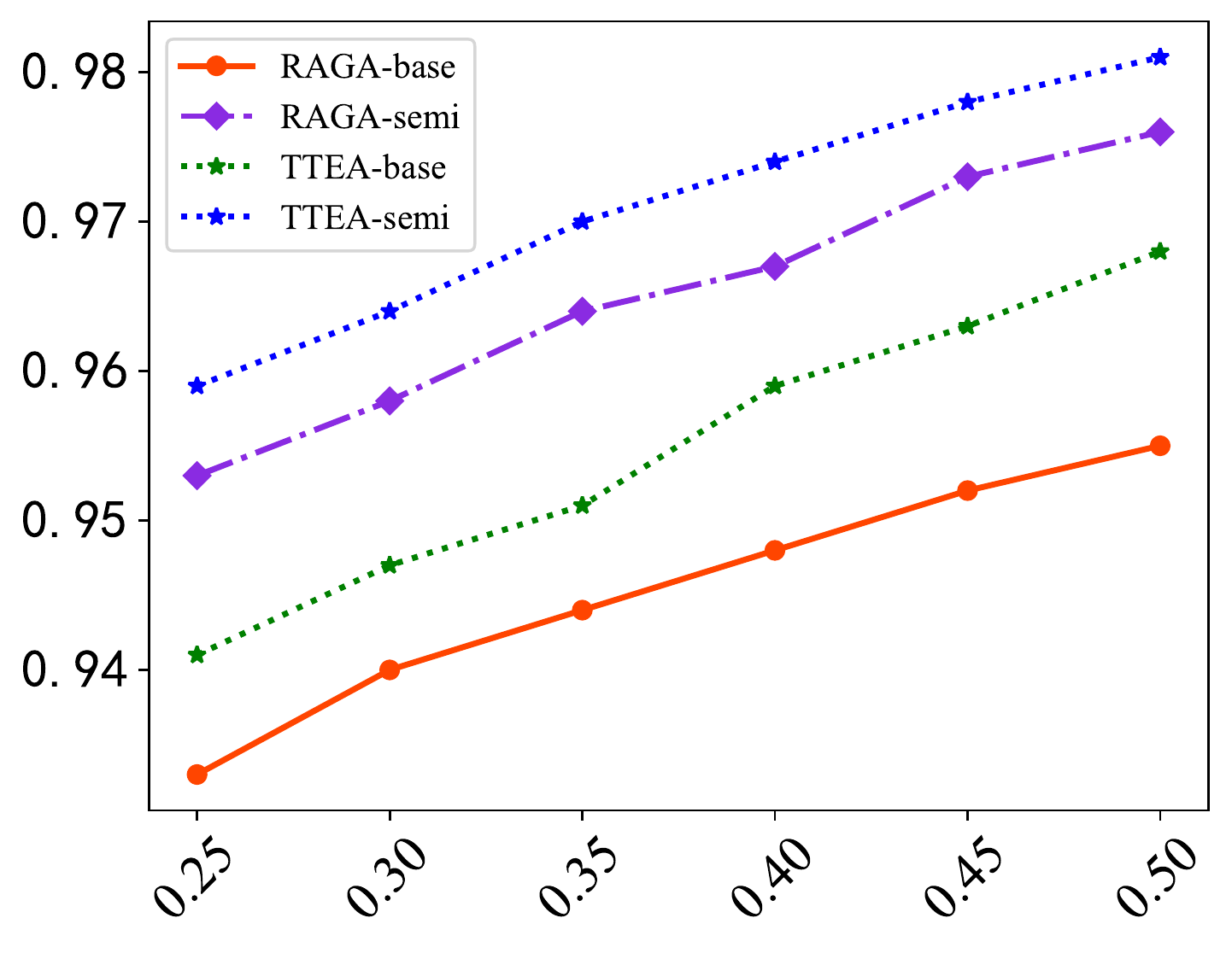}
		\end{minipage}
	}
	\caption{MRR with different training seed pairs.}
	\label{FIG:trainmrr}
\end{figure*}
\subsubsection*{Impact of Seed Entity Pairs}\label{subsubsec6ablation}
To explore the impact of different training seeds, we compare RAGA-l and RAGA-semi with TTEA-base and TTEA-semi by varying the proportion of training seeds from 25\% to 50\% with a step size of 5\%. The results in Figure \ref{FIG:train1}, Figure \ref{FIG:train10} and Figure \ref{FIG:trainmrr} respectively depict Hits@1, Hits@10 and MRR with different seeds proportions. It is showed that TTEA is better than RAGA on both base and semi-supervised local alignment methods two modes for all metrics of three datasets. And as training seeds increase, the Hit@1, Hits@10 and MRR curves of TTEA-base on three datasets are steeper than RAGA-l, which draw the better performance and potentiality.

\section{Related Work}\label{sec2}
Most of the Trans-based methods adopt TransE\citep{bordes_translating_2013} and its variants\citep{transh2014, lin_learning_2015} to embed entity and relation. A line of works embed entity and relation in different latent spaces for different KGs and then construct mapping transformation for EA\citep{chen_multilingual_2017, zhu2017iterative, sun_transedge_2019, qiu_entity_2021, xiang-etal-2021-ontoea}. The second line of works embed entity and relation from different KGs into a unified latent space via parameters sharing\citep{kang_iterative_2020} and extending aligned relation\citep{huang_cross-knowledge-graph_2022}. However, internal correlation of a specific triple is ignored in these methods.

With the application of GNNs on EA, researchers have obtained great improvement by using GNNs. GCN-Align\citep{wang_cross-lingual_2018} is the first work to enhance entity embedding via GCNs, following which GCNs-based methods are extended to aggregate neighbor information. Some works use neighbor entities and relations alignment to tackle EA with the assumption that equivalent entities sharing approximate neighbors\citep{wu_neighborhood_2020}, and some other related efforts utilize topology structure to reinforce entity embedding via GCNs\citep{wu2019jointly, zhu_relation-aware_2020, peng_embedding-based_2020, li_jointly_2022, sun_knowledge_2019}.

And then GATs-based methods are designed considering different neighbor entities contribute different importance. Some works adopt neighbor attention for entity embedding\citep{dual_gated, huang_multi-view_2022, xing_entity_2021, ding_multi_2021, zhu_neighborhood-aware_2019}, in which to our best knowledge, RAGA\citep{zhu_raga_2021} achieves state-of-the-art results by modeling correlation between entity and relation. And others utilize cross-KG attention to spread neighbor information of aligned pairs\citep{wu_relation-aware_2019, yan_soft-self_2021, li_semi-supervised_nodate}.
Moreover, some external resources are integrated into GNNs-based methods to enhance embedding\citep{yang2019aligning, trisedya_entity_2019, yang_cotsae_2020, zhu_cross-lingual_2022, gao_mhgcn_2022}.
Existing GNNs-based methods have effectively improved the performance of EA, but not considered diversity of entities roles and the multi-level representation of ensemble triple.

In the past few years, Boostrapping learning\citep{sun_bootstrapping_2018} and iterative training strategy\citep{zhu2017iterative} are introduced to tackle insufficient seed entity pair. Specifically, bi-directional iterative training strategy\citep{mao_mraea_2020} are widely applied recently\citep{mao_relational_2020, trung_adaptive_2020, zhang_adaptive_2021, qi_multiscale_2022-5}, which is also adopted in TTEA for improving performance.
\section{Conclusion}\label{sec7}
In this paper, to address insufficient utilization of triple specificity and the diversity of entity role, we present a novel framework TTEA -- Type-enhanced Ensemble Triple Representation via Triple-aware attention for Cross-lingual Entity Alignment. By modeling role features and relational interaction between semantic space and type space, TTEA is capable to incorporate ensemble triple specificity and learn cycle co-enhanced head and tail representations. Compared with state-of-the-art baselines, our model achieves the best performance on three real-world cross-lingual datasets.
\section*{Limitations}
Our framework can be free from the limitations of external resources and structural heterogeneity via effectively mining ensemble triple specificity and entity role diversity, which is applicable to most KGs for knowledge completion. However, entity name-based initial embedding adopted by TTEA may not be available, which is a crucial improving part that we will tackle in the future. Moreover, our framework is an element-wise task for cross-lingual event integration, based on which the event level alignment task is another part of our future work.
\section*{Ethics Statement}
Our paper propose TTEA, a novel cross-lingual EA framework modeling triple specificity and role diversity. TTEA neither introduces any social/ethical bias to the model nor amplifies any bias in the data. Our model is built upon public libraries in Pytorch. Moreover, TTEA is trained and tested on public datasets. We do not foresee any direct social consequences or ethical issues.


\bibliographystyle{acl_natbib}
\bibliography{Alignment,custom}

\appendix

\end{CJK}
\end{sloppypar}
\end{document}